\documentclass[11pt]{article}
\usepackage{coling2016/coling2016}
\usepackage{amsmath}
\usepackage{times}
\usepackage{url}
\usepackage{latexsym}
\usepackage[utf8]{inputenc}
\usepackage{graphicx}
\usepackage{color}
\usepackage{textcomp}

\title{
Still not there? Comparing Traditional Sequence-to-Sequence Models to
Encoder-Decoder Neural Networks on Monotone String Translation Tasks
}

\author{Carsten Schnober\textsuperscript{$\dagger\ddagger$}, Steffen Eger\textsuperscript{$\dagger$}, Erik-L\^an Do Dinh\textsuperscript{$\dagger$}, and Iryna Gurevych\textsuperscript{$\dagger\ddagger$}\\[.6em]
   \textsuperscript{$\dagger$}Ubiquitous Knowledge Processing Lab (UKP-TUDA)\\
   Department of Computer Science, Technische Universit\"at Darmstadt\\[.1cm]
   \textsuperscript{$\ddagger$}Ubiquitous Knowledge Processing Lab (UKP-DIPF)\\
   German Institute for Educational Research and Educational Information \\[.1em]
   \url{https://www.ukp.tu-darmstadt.de/}\\
}

\date{}

\begin{document}
\maketitle

\begin{abstract}
We analyze the performance of encoder-decoder neural models and compare them with well-known established methods. The latter represent different classes of traditional approaches that are applied to the monotone sequence-to-sequence tasks OCR post-correction, spelling correction, grapheme-to-phoneme conversion, and lemmatization.
Such tasks are of practical relevance for various higher-level research fields including \textit{digital humanities}, automatic text correction, and speech recognition. 
We investigate how well generic deep-learning approaches adapt to these tasks, and how they perform in comparison with established and more specialized methods, including our own adaptation of pruned CRFs. 
\end{abstract}

\section{Introduction}
\label{sect:introduction}

\blfootnote{
This work is licensed under a Creative Commons Attribution 4.0 International Licence.\\
Licence details: \url{http://creativecommons.org/licenses/by/4.0/}
}

Encoder-decoder neural models \cite{sutskever_sequence_2014} are a
generic deep-learning approach to sequence-to-sequence translation
(Seq2Seq) tasks. They encode an input sequence into a vector
representation from which the 
decoder generates an output.  
These models have shown to achieve state-of-the-art or at least highly
competitive results for various NLP tasks including machine
translation \cite{cho_learning_2014}, conversation modeling  
\cite{vinyals_neural_2015}, question answering
\cite{yin_attention-based_2016}, and, more generally, language
correction \cite{schmaltz_sentence-level_2016,xie_neural_2016}. 
 
We have noticed that, given the enormous interest currently surrounding neural architectures,
recent research appears to somewhat over-enthusiastically praise the
performance of encoder-decoder approaches for Seq2Seq tasks.
For example, while the encoder-decoder G2P model by \newcite{Rao:2015} achieves an extremely low error rate on the CMUdict dataset \cite{kominek_cmu_2004}, the neural architecture itself has a mediocre performance and only outperforms traditional models \emph{in combination} with a weighted finite state transducer. 
Similarly, \newcite{faruqui_morphological_2016} report on ``par or better'' performance of their inflection generation neural architecture.
However, a closer inspection of their results suggests that their system is sometimes worse and sometimes better than traditional approaches. 

Here, we aim for a more {balanced} comparison on three exemplary
monotone\footnote{We call our
  tasks, described below, monotone because relationships between input
  and output 
  sequence characters typically obey monotonicity. That is, unlike in machine
  translation, there are no `crossing edges' in corresponding
  alignments.} Seq2Seq tasks, namely \emph{spelling correction}, \emph{G2P
  conversion}, and \emph{lemmatization}.  
Monotone Seq2Seq tasks
such as morphological analysis/lemmatization,
grapheme-to-phoneme conversion (G2P)
\cite{YaoZ:2015,Rao:2015}, transliteration \cite{Sherif:2007}, and
spelling correction \cite{brill_improved_2000} have been
fundamental problem classes in natural language processing
(NLP) ever since the origins of the field. Their simplicity vis-\`a-vis
non-monotonic problems such as 
machine translation renders them as particularly tractable testbeds of
technological progress. Unlike previous work, which has typically
focussed on only 
one specific subproblem of monotone Seq2Seq tasks at a time, we
consider 
model performances on \emph{three} such tasks simultaneously. 
This leads to a more balanced view on the relative
performance of different models. 

We compare three variants of encoder-decoder models --- including attention-based models  
\cite{bahdanau_neural_2014,luong_effective_2015} and the model proposed by \newcite{faruqui_morphological_2016} --- to three very
well-established baselines for monotone Seq2Seq, namely
Sequitur \cite{bisani_joint-sequence_2008}, DirecTL+ \cite{jiampojamarn_integrating_2010}, and Phonetisaurus \cite{novak_wfst-based_2012}.
We also offer our own contribution\footnote{Our implementation of PCRF-Seq2Seq is available at:\\ \url{https://github.com/UKPLab/coling2016-pcrf-seq2seq}}, which may be considered a variation of
the principles underlying DirecTL+. 
For that purpose, we have adapted higher-order pruned conditional
random fields (PCRFs)
\cite{mueller_efficient_2013,lafferty_conditional_2001} to handle
generic monotone Seq2Seq tasks.   

We find that traditional models appear to still be on par with or better than encoder-decoder models in most cases,
depending on factors such as training data size and the complexity of the task at hand.
We show that neural models unfold their strengths as soon as more complex phenomena need to be learned.
This becomes clearly visible in the comparison between lemmatization and the other tasks we have investigated.
Lemmatization is the only task at hand in which neural models outperform all established systems --- as it is the only one which systematically exhibits long-range dependencies, particularly through Finnish \textit{vowel harmony} (see Section~\ref{sect:results}).
We are thus able to contrast the different challenges imposed by different tasks and show how these differences have significant impact on the performance of encoder-decoder models in comparison to established Seq2Seq models.

To our best knowledge, no systematic comparison with regard to the
suitability of these encoder-decoder neural models for a wider and more generic selection of tasks has been conducted.

\section{Task Description}
\label{sect:tasks}

Throughout, we denote individual tokens in a sequence by
ordinary letters $x$, and a sequence of symbols by 
$\vec{x}$. 
Hence a string of length $s$ is denoted as
$\vec{x}=x_1\dots x_s$.
Real-valued vectors are denoted by bold-faced letters, $\mathbf{x}$. 
 
\textbf{Spelling correction} 
is the problem of converting an `erroneous' input sequence
$\vec{x}$ into a corrected version $\vec{y}$. In terms of errors
committed by humans (typos), spelling correction often deals with errors that
are due to keyboard adjacency of characters
and grapho-phonemic mismatch (e.g.\ 
$emergancy \rightarrow emergency$,
$wuld \rightarrow would$).

\textbf{OCR post-correction} can be seen as a special case of spelling
correction.
OCR (optical character recognition) is the process of
digitizing printed texts automatically, often applied to make text
data from the pre-electronic age digitally available \cite{springmann_ocr_2014}.
Depending on various factors including paper and scan quality, typeface, and OCR engine, OCR error rate can be extraordinarily high \cite{Reynaert:2014}.
OCR post-correction is of particular practical importance in the
field of \emph{digital humanities}.
Here, paper quality, which is often bleached and tainted, and
``unusual'' typefaces typically cause major problems. Unlike in human
spelling correction, OCR errors often arise due to \emph{visual}
similarity of character sub-sequences such as $rn \rightarrow m$
or $li \rightarrow h$. 

Previous works in OCR post-correction apply noisy-channel models \cite{brill_improved_2000} and various extensions \cite{toutanova_pronunciation_2002,cucerzan_spelling_2004,gubanov_improved_2014}, generic string-to-string substitution models \cite{xu_probabilistic_2014}, discriminative models \cite{okazaki_discriminative_2008,farra_generalized_2014}, and user-interactive approaches \cite{reffle_unsupervised_2013}.
Neural network designs including auto-encoders \cite{raaijmakers_deep_2013} and recurrent neural networks \cite{chrupala_normalizing_2014} were also investigated in previous works.

\textbf{G2P} conversion is the problem of converting orthographic
representations into sound representations.
It is the prime example of a monotone Seq2Seq task, which --- 
as a fundamental building block for speech recognition, speech
synthesis, and related tasks --- has been researched for decades. 
It differs from the previous two tasks in that input and
output strings are defined over different alphabets. 

\textbf{Lemmatization} is the task of deriving the lemma from an inflected word form such as \emph{atmest}$\rightarrow$\emph{atmen}. 
The problem is relatively simple for morphologically poor languages like English, but much harder for languages like Finnish. 
The task can be seen as the inverse to 
inflection generation
\cite{Durrett:2013,Ahlberg:2014,Nicolai:2015,faruqui_morphological_2016},
where an 
inflected form is generated from a lemma plus an inflection
tag.  

\section{Data}
\label{sect:data}

Here we detail the data sets used in our experiments; examples are provided in Table~\ref{table:errors}.
These datasets reflect the different Seq2Seq tasks we aim to
investigate.

The \textbf{Text+Berg} corpus \cite{bubenhofer_text+berg-korpus_2015}
contains historic proceedings of the \textit{Schweizer Alpenclub}
(``Swiss Alpine Club'') from the years 1864--1899 in Swiss German and
French. The data has been digitized and OCR errors have been corrected
manually. The corpus contains 19,024 pages, 17,186 of which are in
Swiss German, and 1,838 are in French. 
We have extracted 88,302 unique misrecognized words along with their
manually corrected counterparts.

For our experiments, we have used randomly selected 72K entries for training and test our models on another 9K entries.
Furthermore, we report results for each model trained on a reduced
training set (10K entries). 

\textbf{Twitter Typo Corpus}\footnote{Twitter typo corpus: \url{http://luululu.com/tweet/}}: We use a corpus of 39,172 spelling mistakes extracted from English Tweets with their respective corrections.
The manually corrected mistakes
come with a context word on both sides. 
Again, we have split the data randomly, using a training set of
31K entries and 4K for testing.
We also report results on the same test set when using a
reduced training set with 10K entries. 

The \textbf{Combilex} data set \cite{richmond_robust_2009} provides
mappings from English graphemes to phonetic representations (Table~\ref{table:errors}). 
We use different subsets for training, with 2K, 5K, 10K, and 20K
entries respectively.
Furthermore, we employ a test set with 26,609 entries. 

\begin{table}[h]
\centering
	\begin{tabular}{| r@{}c@{}l l |}
	\hline
		\textbf{P'}reunde & \enspace$\rightarrow$\enspace & \textbf{F}reunde & (misrecognition)\\
		Thal\textbf{ }wand & \enspace$\rightarrow$\enspace & Thal\textbf{}wand & (segmentation)\\
		S\textbf{l}u\textbf{tl}erfi\textbf{m} & \enspace$\rightarrow$\enspace & S\textbf{t}u\textbf{d}erfi\textbf{rn} & (multiple errors)\\
	\hline
		kinaatte & \enspace$\rightarrow$\enspace & kinata & \\
		kinaavat & \enspace$\rightarrow$\enspace & kinata & \\
	\hline
	\end{tabular}
	\begin{tabular}{| r@{}c@{}l | }
	\hline
		to\_\textbf{y}ork\_from & \enspace$\rightarrow$\enspace & to\_\textbf{w}ork\_from\\
		before\_\textbf{t}t\_was & \enspace$\rightarrow$\enspace & before\_\textbf{i}t\_was\\
		with\_my\_daug\textbf{th}er & \enspace$\rightarrow$\enspace & with\_my\_daug\textbf{ht}er\\
	\hline
		Waterloo & \enspace$\rightarrow$\enspace & wOtBr5u\\
		barnacles & \enspace$\rightarrow$\enspace & bArn@k@5z\\
	\hline
	\end{tabular}
	\caption{Training data examples from the four corpora used: OCR detection errors for Text+Berg (top left), Twitter Typo Corpus (top right), Wiktionary Morphology Dataset (Finnish) (bottom left), and Combilex G2P mappings (bottom right).}
	\label{table:errors}
\end{table}

For lemmatization, we use the \textbf{Wiktionary Morphology Dataset} \cite{Durrett:2013}. 
The data set contains inflected forms for different languages and parts of speech, corresponding lemmas, and detailed inflection information, including mood, case, and tense. 
We conduct experiments on the German and Finnish verb datasets and further reduce the size of the latter by considering present tense indicative verb forms in active voice only. 
Note that our results are not comparable to the ones presented by
\newcite{Durrett:2013}, \newcite{Ahlberg:2014}, \newcite{Nicolai:2015}, and \newcite{faruqui_morphological_2016} 
because we focus on lemmatization, not inflection generation, as mentioned. 
We do so because this produces less overhead --- e.g.,
\newcite{faruqui_morphological_2016} train 27 different systems for
German verbs, one for each inflection type --- and there is
a priori not much difference in whether we transform an inflected form
to a lemma or vice versa. 
Hence,  
the relative ordering of the systems we survey should not be affected by this change of direction in the morphological analysis. 
In total, we have used training sets of size 43,929 entries for German verbs and 41,094 entries for Finnish verbs, and dev set and test set sizes of 5,400~(German) and 1,200~(Finnish) entries each.

\section{Model Description}
\label{sect:models}

In this section, we briefly describe encoder-decoder neural models, pruned CRFs, and our three 
baselines. 

\subsection{Encoder-Decoder Neural Models}

We compare three variants of encoder-decoder models: the `classic'
variant and two modifications:

\begin{itemize}
\item
\verb|enc-dec|: Encoder-decoder models using recurrent neural networks
(RNNs) for Seq2Seq tasks were introduced by
\newcite{cho_learning_2014} and \newcite{sutskever_sequence_2014}. 
The encoder reads an input $\vec{x}$ and generates a vector
representation $\mathbf{e}$ from it.
The decoder predicts the output
$\vec{y}$ one 
time step $t$ at a time, based on $\mathbf{e}$.
The probability for each output symbol $y_t$ hence depends on $\mathbf{e}$ and all previously generated output
symbols:
$p(\vec{y} \vert \mathbf{e}) = \prod_{t=1}^{T'} p(y_t \vert
\mathbf{e}, y_1 
\cdots y_{t-1})$ where $T'$ is the length of the output
sequence. 
In NLP, most implementations of encoder-decoder models employ 
LSTM (long short-term memory) layers as hidden units, which extend generic RNN
hidden layers with a memory cell that is able to ``memorize'' and ``forget'' features. 
This addresses the `vanishing gradients' problem and allows to
catch long-range dependencies.

\item
\noindent \verb|attn-enc-dec|: We explore the attention-based
encoder-decoder model proposed by \newcite{bahdanau_neural_2014} (Figure~\ref{fig:attention}). 
It extends the encoder-decoder model by learning to align and
translate jointly.
The essential idea is that the current output unit $y_t$ does not depend on all input units in the same way, as captured by a
`global' vector $\mathbf{e}$ encoding the input. 
Instead, $y_t$ may be conditioned upon local context in the input (to which it pays \emph{attention}).

\begin{figure}[h]
\centering
\includegraphics[scale=0.7]{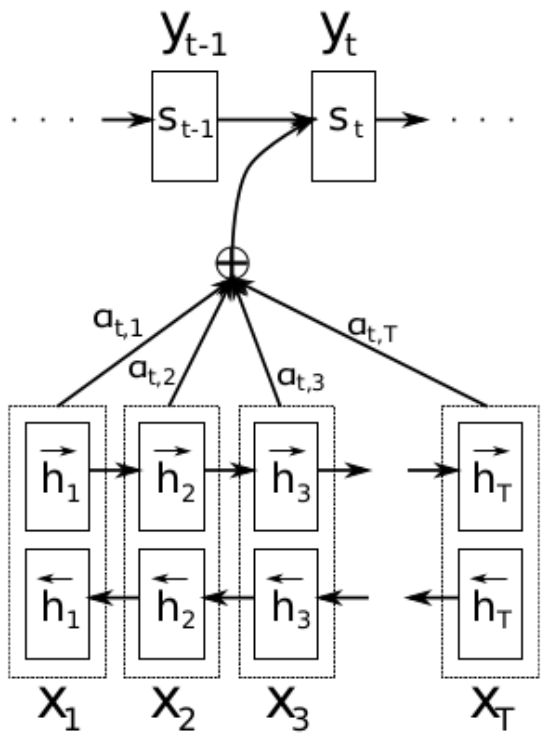}
\caption{In the encoder-decoder model, the encoder (bottom) generates a representation of the input sequence $\vec{x}$ from which the decoder (top) generates the output sequence $\vec{y}$.
The attention-based mechanism (shown here) enables the decoder to ``peek'' into the input at every decoding step through multiple input representations $a_t$.
Illustration from \newcite{bahdanau_neural_2014}.}
\label{fig:attention}
\end{figure}

\item
\verb|morph-trans|: 
\newcite{faruqui_morphological_2016} present a new encoder-decoder
model designed 
for morphological inflection, proposing to feed the input
sequence directly into the decoder. This approach is motivated by the
observation that input and output are usually very similar in 
problems such as 
morphological inflection. Similar ideas have been proposed in
\newcite{Gu:2016} in their so-called ``CopyNet'' encoder-decoder
model (which they apply to text summarization) that allows for
portions of the input sequence to be simply 
copied to the output sequence, without modifications. 
A priori, this observation seems to apply to our tasks too:
at least in spelling correction, the output usually differs only marginally from the input.
\end{itemize}

For the tested neural models, we follow the same overall approach as \newcite{faruqui_morphological_2016}:
we perform decoding and evaluation of the test data using an \textbf{ensemble} of $k=5$ independently trained models in order to deal with the non-convex nature of the
optimization problem of neural networks and the risk of running into a local optimum \cite{collobert_natural_2011}.
The total probability $p_{\text{ens}}$ for
generating an output token $y_t$ is estimated from the individual model output probabilities: $p_{\text{ens}}({y}_t
\vert \cdot) = \dfrac{1}{Z}\prod_{i=1}^k p_i(y_t \vert \cdot)^{\frac{1}{k}}$
with a normalization factor $Z$.

\subsection{Pruned Conditional Random Fields}

Conditional random fields (CRFs) were introduced by
\newcite{lafferty_conditional_2001} and
have been a major workhorse 
for many sequence labeling tasks such as part-of-speech tagging and 
named entity recognition during the 2000s. 
Unfortunately, training and decoding time depend polynomially on the tag
set size and exponentially on the \emph{order} of the CRF. 
Here, order refers to the dependencies on the label side. 
This makes
higher-order CRFs impractical for large training data sizes, which
is the reason why virtually only first-order (linear chain) CRFs
were used until recently.  

\newcite{mueller_efficient_2013} introduced pruned CRFs (PCRFs) that 
approximate the CRF objective function using coarse-to-fine decoding
\cite{charniak_coarse--fine_2005}. PCRFs require much shorter runtime
and are thus able to make use of higher orders. 
Higher orders, in turn, have been shown to be highly beneficial for
coarse and fine-grained part-of-speech tagging, outperforming
first-order models. 

For our tasks, we have adapted the implementation from
\newcite{mueller_efficient_2013} --- originally designed for sequence
labeling --- to general monotone Seq2Seq tasks. Sequence
labeling assumes that an input sequence of length $N$ is mapped to an
output sequence of identical length $N$, while in Seq2Seq tasks, input
string lengths may be shorter, longer, or equal to output string
lengths. 

We address this by 
first \emph{aligning} input and output sequences as
exemplified in 

\begin{center}
\begin{tabular}{ccccccccccc}
  S & l & u & t & l & e & r & f & i & m\\
  S & t & u & d & $\emptyset$ & e & r & f & i & rn
\end{tabular}
\end{center}

\noindent This alignment matches up character subsequences from both
strings. It may include 1-to-zero matches (e.g.\ $l \rightarrow \emptyset$) and 1-to-many
matches (e.g.\ $m \rightarrow rn$). We disallow many-to-1 or many-to-many matches, as 
they cause a problem during decoding: at test time, it is unclear how to segment a new input string into parts with size 
$\ge 1$.
A na\"ive `pipeline' approach (first segment, then translate
the segmented string) leads to error propagation. 
More sophisticated `joint' approaches \cite{jiampojamarn_integrating_2010} are
considerably more computationally expensive.

Once the data is aligned
as above, input and (modified) output sequences are of equal lengths and
we can directly apply higher-order PCRFs. 
Below, we show that orders up to 5 (and possibly beyond) 
are beneficial for the Seq2Seq tasks we
consider.\footnote{In our experiments, higher order CRFs ($> 3$) substantially outperformed first-order models. Typical performance differences were from about 4 to 7\% between first-order and fifth-order models. For brevity, we omit results for orders $\le 3$.} We refer to this model as \textbf{PCRF-Seq2Seq} in the
remainder.

\textbf{Features} Conditional random fields are feature-based, so we
need to decide which features we use. 
In view of the end-to-end nature of neural techniques, requiring little linguistic knowledge, 
we also minimize feature-engineering effort for the traditional approaches and  thus only include very simple features. 
For each position $p$ to tag, we include all consecutive
character 
$m$-grams ($m$ ranges from $1$ to a maximum order of $N$) within a
\emph{window} of size $w$ around $p$; i.e., in total our window covers 
$2w+1$ positions. In our experiments below, we report
results for windows of size $w=4$ and $w=6$. For simplicity, we set $N=w$
in each case. 

\subsection{Further Baseline Systems}
\label{sect:baseline}

Considering the similarity of 
G2P conversion, spelling correction, and lemmatization with regard to their innate monotonicity
\cite{eger_comparison_2016,Nicolai:2015,Eger:2015}, we explore for
all our datasets three further approaches that were originally
designed for G2P conversion. 

\textbf{Sequitur} \cite{bisani_joint-sequence_2008} is a `joint' model for Seq2Seq in the sense of the classic distinction between joint and discriminative models.
Its core  architecture is a model over `joint $n$-grams', also termed `graphones'
in the original publication (that is, pairs of substrings of the $\vec{x}$
and $\vec{y}$ sequence). 

\textbf{DirecTL+} \cite{jiampojamarn_integrating_2010} is a
discriminative model for monotone Seq2Seq that integrates joint
$n$-gram features. It jointly learns input segmentation,
output prediction, and sequence modeling.  
Since it is based on ordinary CRFs, it is virtually impossible to use
this system with higher orders for all practically relevant datasets due to very long training times.
Moreover, the system is generally very slow because it jointly learns to segment and translate, as mentioned. For this reason, we have only 
tested 
it on the Combilex dataset (Table~\ref{table:g2p_accuracy}), run with
comparable parametrizations (context size, etc.) as PCRF-Seq2Seq.

\textbf{Phonetisaurus} \cite{novak_wfst-based_2012} implements a weighted finite state transducer (WFST) to align input and output tokens.
The EM-driven algorithm is capable of learning multiple-to-multiple alignments where we restrict both sides to a maximum of 2.
The alignments learned from the training data are subsequently used to
train a character-based $n$-gram language model.
For brevity, we only report results for models with $n=8$, which outperformed lower-order models in our experiments.

\section{Results and Analysis}
\label{sect:results}

\subsection{Model Performances}
We report the results of all our experiments in terms of \emph{word accuracy} (WAC), i.e., the fraction of completely correctly predicted output sequences. 
Table~\ref{table:correction_accuracy} lists WACs for all our systems on the OCR post-correction task (\textit{Text+Berg}, full and reduced training set)  
and on the spelling correction task (Twitter, full and reduced training set).
Table~\ref{table:g2p_accuracy} reports WAC of all tested models on the Combilex dataset
with models trained on training sets of different sizes.
Table~\ref{table:lemmas_accuracy} reports WAC for the lemmatization task on the morphology dataset.

\begin{table}[tb]
\begin{center}
\begin{tabular}{| l | c | c || c | c | c | }
\hline & \textbf{Text+Berg \small (72K)} & \textbf{Text+Berg \small (10K)} & \textbf{Twitter \small (31K)} & \textbf{Twitter \small (10K)} \\
\hline
attn-enc-dec \small (1 layer, size 100) & 66.80\% & 61.71\% & 66.25\% & 60.99\% \\ 
attn-enc-dec \small (1 layer, size 200) & 68.29\% & \textbf{\underline{63.00\%}} & 67.81\% & 59.36\% \\ 
attn-enc-dec \small (2 layers, size 100) & 68.30\% & 62.27\% & 69.29\% & \underline{63.31\%} \\ 
attn-enc-dec \small (2 layers, size 200) & \underline{69.74\%} & 62.87\% & \underline{69.70\%} & 62.01\% \\ 
\hline
enc-dec \small (1 layer, size 100) & 50.96\% & 39.99\% & 60.91\% & 52.90\% \\ 
enc-dec \small (1 layer, size 200) & 53.65\% & 41.52\% & 63.39\% & 55.76\%\\ 
enc-dec \small (2 layers, size 100) & 56.94\% & 42.53\% & \underline{65.94\%} & \underline{56.50\%} \\ 
enc-dec \small (2 layers, size 200) & \underline{59.01\%} & \underline{46.01\%} & 63.70\% & 53.74\% \\ 
\hline
morph-trans \small (1 layer, size 100) & 55.96\% & 49.37\% & 41.46\% & 39.19\% \\ 
morph-trans \small (1 layer, size 200) & 54.22\% & \underline{49.63\%} & 36.28\% & 30.74\% \\ 
morph-trans \small (2 layers, size 100) & \underline{56.11\%} & 47.35\% & \underline{44.42\%} & \underline{42.02\%} \\ 
morph-trans \small (2 layers, size 200) & 49.27\% & 45.55\% & 30.33\% & 29.61\% \\ 
\hline
PCRF-Seq2Seq \small (order 4, $w=$ 6) & \textbf{\underline{74.67\%}} & 62.24\% & 73.52\% & 59.97\% \\ 
PCRF-Seq2Seq \small (order 5, $w=$ 6) & 74.22\% & 62.47\% & 74.03\% & 60.19\% \\ 
PCRF-Seq2Seq \small (order 4, $w=$ 4) & 74.55\% & \underline{62.75\%} & \underline{\textbf{74.87\%}} & \textbf{\underline{63.59\%}} \\
\hline Phonetisaurus \small ($n=8$) & 60.89\% & 51.84\% & 69.52\% & 55.76\% \\ 
\hline Sequitur & 68.04\% & 57.30\% & 70.74\% & 58.90\% \\ 
\hline 
\end{tabular}
\end{center}
\caption{Word accuracies (WACs) for all encoder-decoder models, 
PCRF-Seq2Seq,
and baselines for the \textbf{OCR post-correction} task and for the \textbf{spelling correction} task.
Best configurations for each model are underlined, overall best results are bold-faced.} 

\label{table:correction_accuracy}
\end{table}

\begin{table}[tb]
\begin{center}
\begin{tabular}{| l | c | c | c | c | }
\hline
 & \textbf{Combilex \small (20K)} & \textbf{Combilex \small (10K)} & \textbf{Combilex \small (5K)} & \textbf{Combilex \small (2K)} \\
\hline
attn-enc-dec \small (1 layer, size 100) & 57.39\% & 54.40\% & 41.19\% & 35.68\% \\ 
attn-enc-dec \small (1 layer, size 200) & 61.86\% & 57.92\% & 46.72\% & 38.31\% \\ 
attn-enc-dec \small (2 layers, size 100) & 66.74\% & \underline{60.52\%} & \underline{55.89\%} & 44.13\% \\ 
attn-enc-dec \small (2 layers, size 200) & \underline{67.36\%} & 59.62\% &	55.07\% & \underline{44.26\%} \\ 
\hline
enc-dec \small (1 layer, size 100) & 54.03\% &	48.25\% & 36.62\% &	18.17\%  \\ 
enc-dec \small (1 layer, size 200) & 55.27\% &	49.81\% & 36.19\% &	\underline{18.41\%}  \\ 
enc-dec \small (2 layers, size 100) & \underline{57.77\%} &	\underline{51.91\%} & 38.97\% & 16.68\%  \\ 
enc-dec \small (2 layers, size 200) & 56.95\% &	50.69\% & \underline{39.01\%} &	17.83\%  \\ 
\hline
morph-trans \small (1 layer, size 100) & 48.82\% &	\underline{43.30\%} & \underline{33.63\%} & \underline{18.97\%} \\ 
morph-trans \small (1 layer, size 200) & \underline{49.72\%} & 43.15\% & 32.42\% & 18.76\% \\ 
morph-trans \small (2 layers, size 100) & 49.58\% &	42.05\% & 28.69\% & 15.13\% \\ 
morph-trans \small (2 layers, size 200) & 44.36\% & 35.14\% & 23.08\% & 12.74\% \\ 
\hline
PCRF-Seq2Seq \small (order 4, $w=$ 6) & 72.14\% & 64.39\% &	55.66\% & 42.82\% \\ 
PCRF-Seq2Seq \small (order 5, $w=$ 6) & \underline{72.23\%} &	64.32\% &	55.58\% &	42.62\% \\
PCRF-Seq2Seq \small (order 4, $w=$ 4) & 71.74\% & \underline{64.71\%} & \textbf{\underline{56.89\%}} & \textbf{\underline{44.74\%}} \\ 
\hline DirecTL+  & 72.23\% &	\textbf{65.09\%} &	55.75\% &	42.95\% \\ 
\hline Phonetisaurus \small ($n=8$) & \textbf{72.29\%} &	64.14\% &	55.28\% &	42.21\%  \\ 
\hline Sequitur  & 70.57\% &	62.57\% &	54.03\% &	41.94\% \\ 
\hline 

\end{tabular}
\end{center}
\caption{WACs for all encoder-decoder models, PCRF-Seq2Seq,
and baselines for the \textbf{G2P} task.
Best configurations for each model are underlined, overall best results are bold-faced.}

\label{table:g2p_accuracy}
\end{table}

\begin{table}[tb]
\begin{center}
\begin{tabular}{| l | c | c | }
\hline & \textbf{German Verbs (44K)} & \textbf{Finnish Verbs (41K)} \\
\hline
attn-enc-dec \small (1 layer, size 100) & 93.67\% & \textbf{\underline{98.00\%}} \\ 
attn-enc-dec \small (1 layer, size 200) & 93.17\% & 96.92\% \\ 
attn-enc-dec \small (2 layers, size 200) & 92.39\% & 97.08\% \\ 
attn-enc-dec \small (2 layers, size 200) & \textbf{\underline{94.83\%}} & 96.42\% \\ 
\hline
enc-dec \small (1 layer, size 100) & 77.31\% & 94.83\% \\ 
enc-dec \small (1 layer, size 200) & \underline{82.00\%} & 94.67\% \\ 
enc-dec \small (2 layers, size 200) & 79.50\% & \underline{95.67\%} \\ 
enc-dec \small (2 layers, size 200) & 76.30\% & 95.58\% \\ 
\hline
morph-trans \small (1 layer, size 100) & 91.89\% & 96.17\% \\ 
morph-trans \small (1 layer, size 200) & 93.24\% & 96.75\% \\ 
morph-trans \small (2 layers, size 200) & 93.02\% & \underline{97.08\%} \\ 
morph-trans \small (2 layers, size 200) & \underline{93.63\%} & 96.75\% \\ 
\hline
PCRF-Seq2Seq \small (order 4, $w=$ 6) & \underline{94.22\%} & \underline{94.08\%} \\ 
PCRF-Seq2Seq \small (order 5, $w=$ 6) & 93.77\% & 94.00\% \\ 
PCRF-Seq2Seq \small (order 4, $w=$ 4) & 93.44\%	& 93.33\% \\
\hline Phonetisaurus \small ($n=8$) & 86.62\% & 93.42\% \\ 
\hline Sequitur & 85.63\% & 92.92\% \\ 
\hline 
\end{tabular}
\end{center}
\caption{WACs for all encoder-decoder models, PCRF-Seq2Seq,
and baselines for the \textbf{lemmatization task}.
Best configurations for each model are underlined, overall best results are bold-faced.}

\label{table:lemmas_accuracy}
\end{table}

For the encoder-decoder models, we report the results with one and two layers of sizes 100 and 200 each. 
We have additionally conducted sample experiments with larger networks which have shown that neither increasing the number of layers nor the size of the layers leads to further improvements.
For the PCRF-Seq2Seq models, we report results for windows of sizes $w=4$ and $w=6$. 
We note that, a priori, 
more training data tends to favor larger context size $w$, whereas a
large $w$ may lead to overfitting when training data is small. The
same holds for model order.

While more training data obviously increases WAC for every model, the specific impact varies. 
In general, (attention-based) encoder-decoder models deal relatively
well with limited test data in our experiments, achieving WACs
comparable to PCRF-Seq2Seq.  
In contrast, they appear to benefit less from increasing data sizes than CRFs do. 
On the Twitter dataset, for instance, the best-performing
encoder-decoder model increases WAC by 7.7 percentage points when
tripling the training data size. 
At the same time, the 5th-order PCRF-Seq2Seq WAC increases by 13.8.
When a large amount of training data is available, CRFs therefore
consistently outperform neural models, and so do the specialized
baseline systems on the G2P conversion tasks
(Table~\ref{table:g2p_accuracy}). 

Summarizing, we find that PCRF-Seq2Seq performs best among the tested
systems for the two spelling correction tasks when large training data is available.
The best performance of PCRF-Seq2Seq is roughly 6-7~percentage points better than
the best performance of an encoder-decoder model for both Twitter\,31K
and Text$+$Berg\,72K. For small training set sizes, PCRF-Seq2Seq and
the encoder-decoder models are on
a similar level. 
For the G2P task, an
analogous pattern emerges. Moreover, here, all classical systems
appear to perform similarly, with DirecTL$+$ and PCRF-Seq2Seq
marginally outperforming the others. 
For lemmatization, the overall picture looks different.
For Finnish verbs, we observe the only case in which attention-based encoder-decoder systems clearly outperform all other approaches. 
For German, neural models also achieve the best results, albeit only marginally above PCRF-Seq2Seq.

Previous works that employed encoder-decoder models successfully focused on tasks like machine translation and grammar correction in which more challenging linguistic phenomena such as long-range dependencies and `crossing edges` (re-ordering) occur frequently.
In our experiments, too, neural models only outperform traditional ones when long-range dependencies become relevant, namely in lemmatization.
In all other tasks at hand, in contrast, neural models perform worse or equal.

The afore-mentioned, more complex linguistic phenomena intuitively require a more global view on
long input sequences which is hard to impossible to model for
approaches that cannot look beyond a statically defined context. 
Spelling mistakes, OCR errors, and G2P, however, 
largely depend on a very local context.
For instance, OCR systems typically do not consider more than a small context when estimating the probability of a character. Regarding the G2P task, phonetics is generally independent of characters that occur more than two or three positions before of after, at least in most cases and in English. 
The same is presumably true for human typos, where a mistaken key stroke may be the result of a previous key's position, but does not correlate to any key that was hit several time steps before.
Hence, neural networks are unable to benefit from their often advantageous capability of modeling long-range dependencies here.

Especially the Finnish lemmatization experiments confirm that the capability of dealing with long-range dependencies plays an important role.
Finnish \textit{vowel harmony} makes a vowel control other vowels in the word, potentially across multiple syllables; see \newcite{faruqui_morphological_2016} for more detailed explanation. 

As a side note, our results are in line with the common notion that the specific impact of a neural network's size (number and sizes of layers) is almost unpredictable.
Our results can only confirm the general rule-of-thumb that larger networks are better for larger training sets, while models with fewer parameters outperform larger ones when training data is smaller.

\subsection{Training Time}

Another potentially limiting factor for the applicability of a model in real-world scenarios, especially for large datasets, is training time.
Under all circumstances, weighted finite state transducers (Phonetisaurus) are trained  by magnitudes faster than all other approaches.
Training times range between as little as
6~seconds for the smallest training set and 247~seconds for the full \textit{Text+Berg} training set (72K entries).

In comparison, training times for the encoder-decoder models range from 2 to 80 hours (without using GPUs) for 30 epochs, depending on the sizes of the networks and the training data.
Furthermore, there is no noticeable difference between either of the three encoder-decoder variations.
Training time increases approximately linearly with the number of layers, the size of the layers, and the training data size. All these factors add up, meaning that doubling both the number of layers and the size of the layers approximately quadruples training time.

Contrasting DirecTL+ with PCRF-Seq2Seq, both of which rest on similar principles and also perform similarly in our experiments on the G2P task, we find that training PCRF-Seq2Seq was a factor of 30 or 50 times faster than DirecTL+ on Combilex (2K) and Combilex (5K), respectively. In general, training for PCRF-Seq2Seq across our datasets was in the order of minutes to (few) hours.

\subsection{Error Analysis}
\label{sect:analysis}

\begin{figure}[h]
\centering
\includegraphics[width=\textwidth]{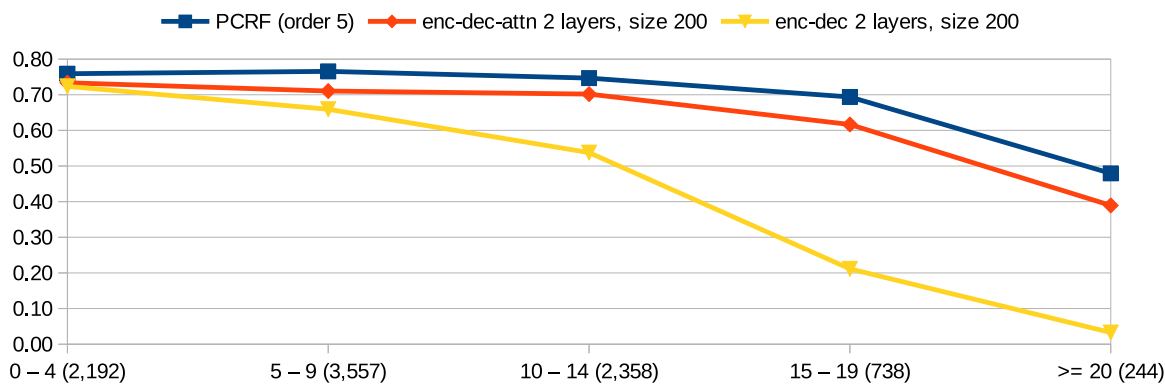}
\caption{WAC of PCRF-Seq2Seq and encoder-decoder neural models with and without attention-based mechanisms as a function of input string length (number of training samples) on \textit{Text+Berg} \textbf{OCR post-correction}.}
\label{fig:textberg_lengths}
\end{figure}

\begin{figure}[h]
\centering
\includegraphics[width=\textwidth]{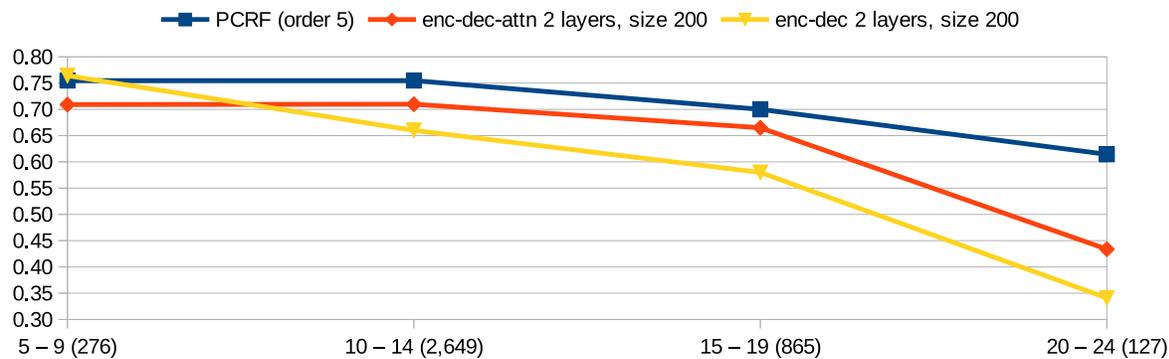}
\caption{WAC of PCRF-Seq2Seq and encoder-decoder neural models with and without attention-based mechanisms as a function of input string length (number of training samples) on the Twitter \textbf{spelling correction} task.} 
\label{fig:twitter_lengths}
\end{figure}

We divided three of our test sets (\textit{Text+Berg}, Twitter, and
Combilex) by input string lengths and evaluated PCRF-Seq2Seq and
encoder-decoder neural models on these subsets of the test data. 
As illustrated in Figures~\ref{fig:textberg_lengths} and
\ref{fig:twitter_lengths},
we observe a consistent tendency:
PCRF-Seq2Seq performs relatively robustly over input strings of
different lengths, while the performance of the encoder-decoder models
plummets more drastically with sequences becoming longer, in particular
those without attention-mechanism.

For shorter sequences, we observe that standard encoder-decoder models even slightly outperform their attention-based counterparts as well as PCRF-Seq2Seq on both the Twitter spelling correction task (Figure~\ref{fig:twitter_lengths}) and on G2P conversion, 
in contrast to their rather low performance on the full datasets. 
On the \textit{Text+Berg} data, all systems achieve approximately equal WAC for short sequences (Figure~\ref{fig:textberg_lengths}).

For longer sequences, the performance of the encoder-decoder models drops dramatically on all data sets.
This effect is also visible, albeit less strong, for the attention-based variant.
This can be seen particularly well on the OCR post-correction task (Figure~\ref{fig:textberg_lengths}), where the test set contains numerous long sequences:
the accuracy rate for the standard encoder-decoder model drops from 73.32\% on very short sequences to below 10\% for very long ones ($\geq 20$), whereas the attention-based model drops less drastically to 38.93\% (from 76.92\%).
At the same time, PCRF-Seq2Seq behaves more stably, 
particularly on the Twitter data (Figure~\ref{fig:twitter_lengths}).
For the Combilex data, the picture looks very similar ---
we omit these results for brevity. 

\section{Conclusions}
\label{sect:conclusion}

The 
generality of neural networks makes them appealing for a wide range of possible tasks. In the scope of this work, we have applied encoder-decoder neural models to monotone Seq2Seq tasks.  
We have shown that they 
can perform comparably to more specialized models in
some cases, but cannot (yet) consistently 
outperform established approaches, and are sometimes still
substantially below them.
Furthermore, the advantage of having rendered feature engineering and hyper-parameter optimization in the traditional sense unnecessary is notoriously substituted by the search for optimal neural network topologies.

At first sight, our analyses based on string lengths
are in line with those reported by \newcite{bahdanau_neural_2014}. They state that --- for the field of machine translation --- the attention mechanism leads to improvements over the standard encoder-decoder model on longer sentences.
We also observe this positive impact for our tasks, where the attention-based mechanism alleviates the drastic performance drop of the standard encoder-decoder models on long sequences to some extent.
At the same time, we see that very performance drop persisting --- CRFs still outperform encoder-decoder models on long sequences, even when employing attention-mechanisms.
As described in Section~\ref{sect:analysis}, neural models are only able to successfully compete when more complex phenomena occur, on which traditional models fail.
Nevertheless, previous works such as \newcite{vukotic_is_2015} also indicate that even in more complex sequence labeling tasks such as spoken language understanding, neural networks are not guaranteed to outperform CRFs. 

The task-specific extensions to the encoder-decoder proposed by
\newcite{faruqui_morphological_2016} have been shown to produce mostly bad results in our settings.
This is particularly surprising for the OCR data, for which 
input and output sequences are usually very similar, so that we had
expected that re-feeding the input to the decoder should be equally
beneficial in that domain. As discussed, one explanation might be that OCR, or
spelling correction generally, putatively exhibits few long-range dependencies. 
This might explain why 
the \verb|morph-trans| approach works quite well and competitive
in morphological analysis 
tasks, as re-confirmed in our experiments. Thus, long-range
dependencies might actually be a more crucial aspect for the performance of the model presented by \newcite{faruqui_morphological_2016}
than the similarity between input and output sequence. 

We conclude that neural networks are far from completely replacing established methods at this point, as the latter can be both faster and more accurate, depending on the properties of the task at hand.
A systematic analysis of the complexities and challenges a particular task imposes, remains unavoidable. 
At the same time, one can argue that encoder-decoder neural models are a relatively recent development and might continue to improve much over the next years. 
Being very generic and largely task-agnostic, they are already able to
outperform traditional and specialized approaches under certain
circumstances.

\section*{Acknowledgements}

This work has been supported by the Volkswagen Foundation as part of the Lichtenberg-Professorship Program under grant {\fontfamily{cmr}\selectfont\textnumero} I/82806, and by the German Institute for Educational Research (DIPF), as part of the graduate program ''Knowledge Discovery in Scientific Literature`` (KDSL).

\bibliographystyle{coling2016/acl}
\bibliography{coling2016}

\end{document}